\documentclass[runningheads]{llncs}
\usepackage[T1]{fontenc}
\usepackage{biblatex}   
\usepackage{amsmath}
\usepackage{graphicx}
\usepackage{xcolor}
\usepackage{float}
\usepackage{caption}
\usepackage{subcaption}

\usepackage{booktabs}
\usepackage{multirow}
\usepackage[titletoc]{appendix}
\usepackage{pifont}

\usepackage{multirow}
\usepackage{hyperref}

\bibliography{main.bib}

\begin{document}
\title{EfficientPose 6D: Scalable and Efficient 6D Object Pose Estimation}
\titlerunning{Scalable and Efficient 6D Object Pose Estimation}
%
\author{Zixuan Fang\orcidID{0009-0003-1926-8023},
        Thomas P\"ollabauer\orcidID{0000-0003-0075-1181},
        Tristan Wirth\orcidID{0000-0002-2445-9081},
        Sarah Berkei\orcidID{0000-0002-7986-1414},
        Volker Knauthe\orcidID{0000-0001-6993-5099},
        Arjan Kuijper\orcidID{0000-0002-6413-0061}}

\maketitle              
\begin{abstract}
In industrial applications requiring real-time feedback, such as quality control and robotic manipulation, the demand for high-speed and accurate pose estimation remains critical.
Despite advances improving speed and accuracy in pose estimation, finding a balance between computational efficiency and accuracy poses significant challenges in dynamic environments.
Most current algorithms lack scalability in estimation time, especially for diverse datasets, and the state-of-the-art (SOTA) methods are often too slow.
This study focuses on developing a fast and scalable set of pose estimators based on GDRNPP to meet or exceed current benchmarks in accuracy and robustness, particularly addressing the efficiency-accuracy trade-off essential in real-time scenarios.
We propose the AMIS algorithm to tailor the utilized model according to an application-specific trade-off between inference time and accuracy.
We further show the effectiveness of the AMIS-based model choice on four prominent benchmark datasets (LM-O, YCB-V, T-LESS, and ITODD).

\keywords{Monocular 6D pose estimation  \and efficient \and fast}
\end{abstract}

\section{Introduction}
In the field of computer vision, object pose estimation can be considered one of the most important tasks.
It aims to determine the rotation and translation of an object in three dimensions respectively.
Due to its wide range of applications such as robotics, autonomous driving, and augmented reality, researchers have attempted to tackle this problem.
In robotics, for example, high-quality 6D pose estimation plays a crucial role in autonomous robotic object manipulation in real-world scenarios such as picking up industrial bins \cite{deng2020self,}.\\
In applications requiring real-time feedback, low inference times play an important role in pose estimation. During robot navigation or autonomous driving, pose estimation systems must be capable of analyzing and responding rapidly and accurately to changes in the environment in order to prevent collisions and ensure the safety of the robot or passenger.
Fast pose estimation can be used for sports analysis and video surveillance to capture critical motions and behaviors, which can be utilized for real-time decision support, as well.
As a result, enhancing the processing speed of a pose estimation algorithm is essential to broaden its practical applications.\\
Often the inference time of a model is closely linked to its accuracy.
Based on this observation, there are a lot of applications that have strong constraints regarding their inference time budget.
For these applications, finding the most accurate model, that fulfils its inference time constraints is key.
Tan et al. \cite{tan2020efficientdet} introduce a family of algorithms for Object Detection, that perform well under different time budgets, giving the user the possibility to chose the variation, that achieves the highest accuracy given their time budget.
Pöllabauer et al. \cite{pollabauer2024fast} propose such a family of architectures for 6D pose estimation.\\
In this work, we propose multiple candidate architectures based on GDRNPP \cite{liuShanicelGdrnpp_bop20222024}, which constitutes an enhanced version of GDR-Net \cite{wang2021gdr}, that optimize inference time while maintaining or surpassing its accuracy on multiple BOP challenge \cite{BOPBenchmark6Da} datasets.
Furthermore, we introduce the Adaptive Margin-Dependent Iterative Selection (AMIS) algorithm that selects a subset out of candidate architectures, that constitute a beneficial trade-off between inference time and accuracy over multiple datasets.
The proposed AMIS algorithm can be applied to a diverse range of task-specific datasets, allowing the choice of a model that reflects the domain-specific requirements with regard to a trade-off between inference time and accuracy.

In summary,
\begin{itemize}
    \item we propose 40 candidate architectures based on modifications of GDRNPP \cite{liuShanicelGdrnpp_bop20222024} by adapting backbone and Geo Head architecture with the primary goal to enhance the resulting inference time while maintaining high accuracy,
    \item we present the AMIS algorithm, which identifies a suitable set of candidate models that constitute an optimal trade-off between inference time and their 6D pose estimation quality over multiple datasets, and
    \item we present quantitative results of the candidate models identified by the proposed AMIS algorithm for the LM-O, YCB-V, T-LESS, and ITODD datasets \cite{xiang2017posecnn, drost2017introducing, hodan2017t,}.
\end{itemize}
\section{Related Work}
\subsection{6D Object pose estimation}

The estimation of 6D positions from monocular RGB-D pictures has a multitude of applications in the industry, e.g., robot grasping, and therefore constitutes an important computer vision task overall.
Especially in the industrial area the need for real-time 6D object pose estimation is prevalent \cite{gorschluter2022survey}, therefore this work focuses on high accuracy applications under constraint inference time.\\
Hodan et al. \cite{hodan2024bop} give a comprehensive overview over the recent developments in 6D object pose estimation as benchmarked by the BOP challenge, that uses a wide variety of datasets and relevant metrics.
Li et al. \cite{li2019cdpn} increase accuracy and robustness by introducing Coordinates-based Disentangled Pose Network (CDPN) that uniquely separates the prediction of rotation from the prediction of translation, demonstrating a high level of flexibility and efficiency even with texture-less and occluded objects.
Labbe et al. \cite{labbe2020cosypose} propose Cosypose, which is able to estimate the 6D pose of multiple objects in scenes captured based on unknown camera viewpoints. From individual images, hypotheses of object poses are generated, which are matched  across different views to estimate both the camera viewpoints and the poses in a unified scene framework.
By employing an object-level bundle adjustment, the method innovatively manages object symmetries without requiring depth information, improving the accuracy and robustness  by minimizing reprojection errors in complex multi-object, multi-view scenarios.
Wang et al. \cite{wang2021gdr} propose GDR-Net, introducing a dense correspondence-based intermediate geometric representations, which in contrast to earlier strategies that set up 2D-3D correspondences and subsequently apply PnP/RANSAC strategies, allows for end-to-end training. As a result of this approach, the 6D pose can be directly regressed, combining the advantages of both direct and indirect methods.
GDRNPP \cite{liuShanicelGdrnpp_bop20222024} upgrades GDR-Net by introducing stronger domain randomization operations, such as background replacement and color enhancements, and most prominently replacing the ResNet-34 backbone with ConvNeXt.
We discuss the architecture of GDR-Net and GDRNPP in more detail in section \ref{sec:preliminaries}.
Haugaard and Buch \cite{haugaard2022surfemb} propose an unsupervised approach, that is employed to learn dense, continuous 2D-3D correspondence distributions on object surfaces without prior knowledge of visual ambiguities such as symmetry. They utilize a compact, fully connected key model and an encoder-decoder query model, both operating within object-specific latent spaces.
Zebrapose \cite{su2022zebrapose} introduces a discrete descriptor that provides a more detailed and accurate mapping of the object surface as compared to previous methods, which allows encoding the surface of an object more efficiently by incorporating a hierarchical binary grouping. Furthermore, they propose a novel coarse-to-fine training strategy that enhances the accuracy by enabling fine-grained correspondence prediction.

\subsection{Speed and Efficiency in Deep Learning}
Modern deep learning architectures have been specifically tailored to address the computational bottlenecks in 6D pose estimation. For instance, lightweight neural networks that incorporate depthwise separable convolutions, such as MobileNet and EfficientNet, have demonstrated significant reductions in computational complexity and latency without compromising accuracy by reducing the number of parameters and operations, thereby speeding up the inference time and allowing for scaling and using different configurations \cite{liang2021efficient, wang2024lightweight,}.\\
Beyond architectural changes, algorithmic adjustments also play a crucial role. For example, employing more sophisticated loss functions that focus on critical parts of the pose estimation task or the application of better regularization techniques can lead to more efficient learning dynamics and prevent overfitting. \cite{liu2023linear, gonzalez2020effective}.\\
Pöllabauer et al. \cite{pollabauer2024fast} propose a set of scalable 6d pose estimation architectures over a wide scale of inference time budgets based on a fixed set of datasets from the BOP challenge \cite{BOPBenchmark6Da}. In contrast to that, we propose a strategy that is able to automatically identify architectures with a beneficial trade-off between inference time and accuracy for arbitrary 6D pose estimation datasets.
\begin{figure}[H]
    \centering
    \includegraphics[width=\textwidth,keepaspectratio]{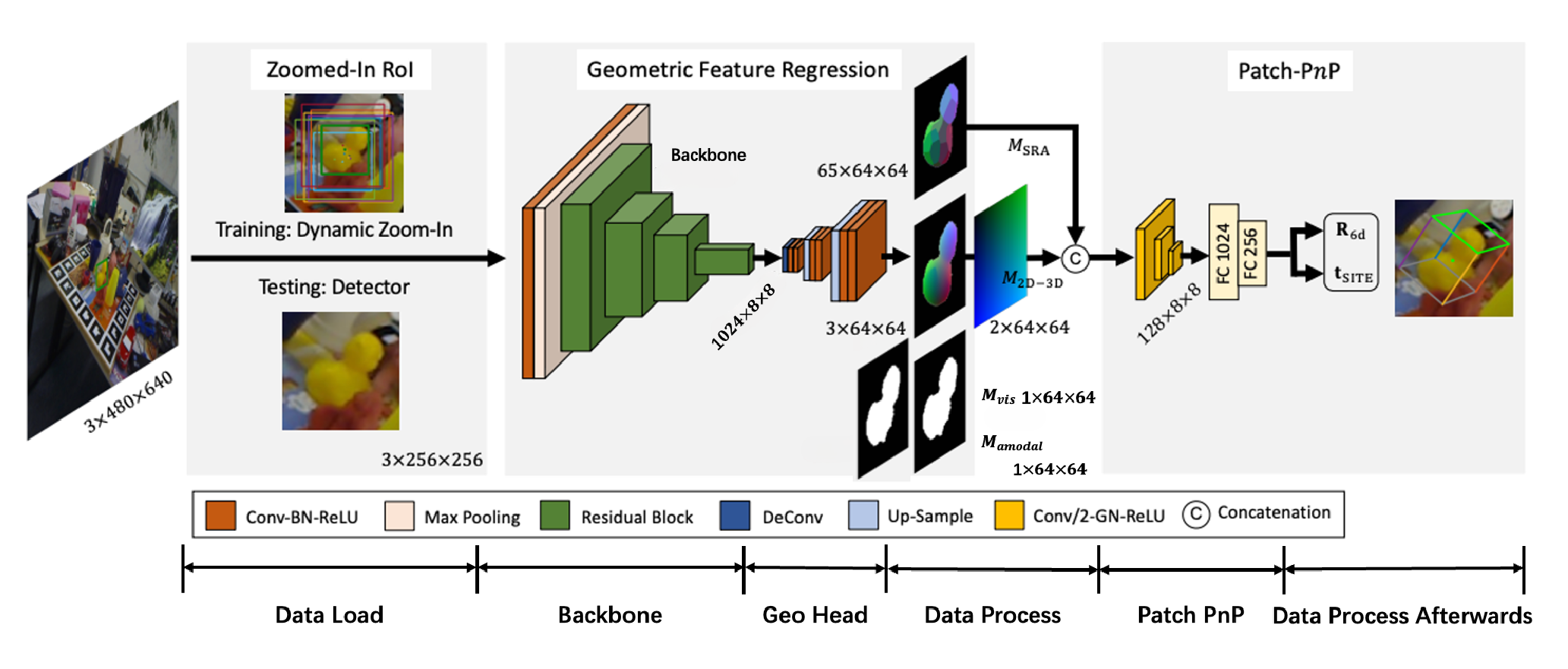}
    \caption{Architecture of GDRNPP / GDR-Net \cite{wang2021gdr}}
    \label{fig:Architecture of GDRNPP}
\end{figure}

\section{Preliminaries}
\label{sec:preliminaries}
\begin{figure}[htbp]
    \centering
    \includegraphics[width=0.6\textwidth,keepaspectratio]{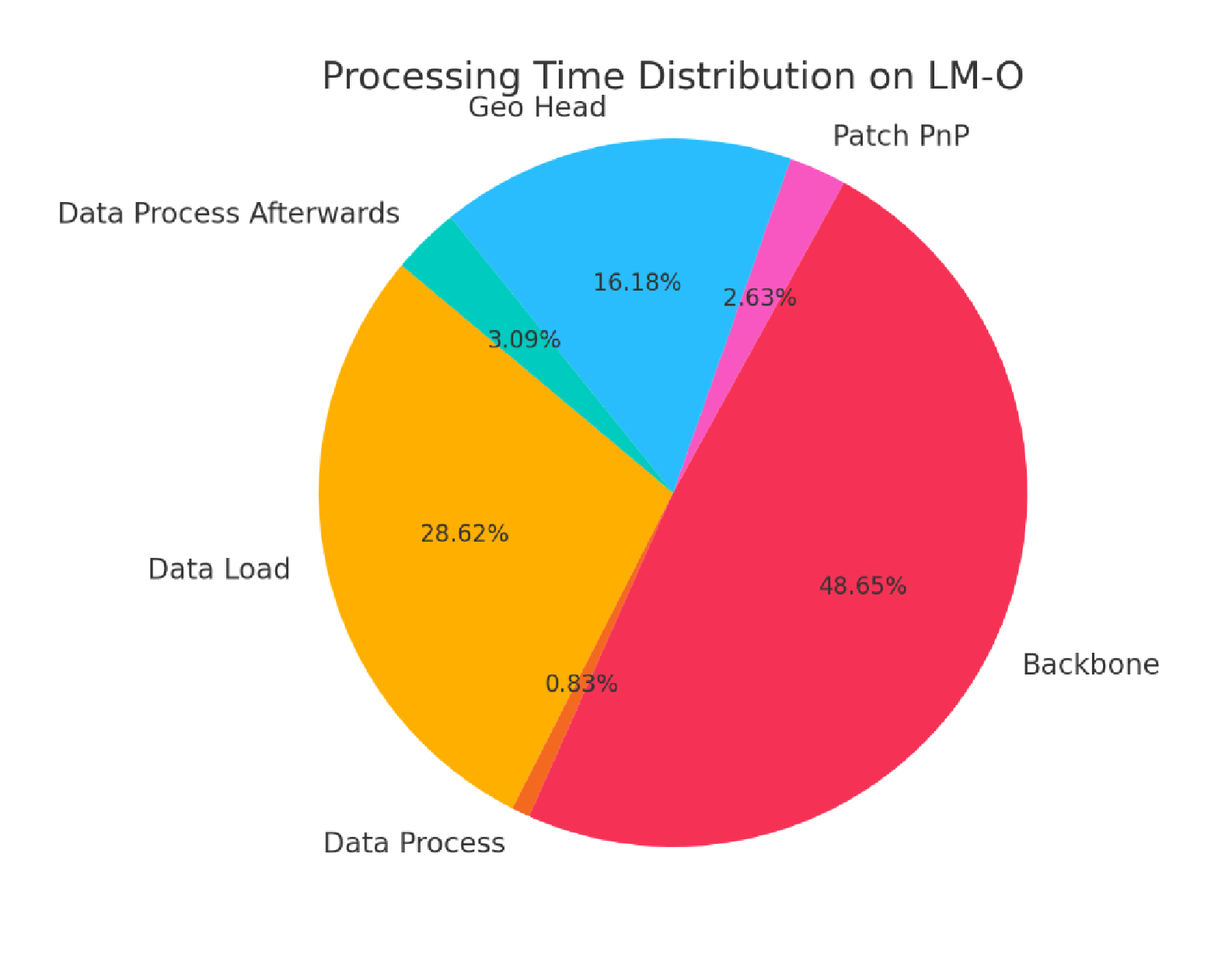}
    \caption{Relative required runtime of the six conceptual stages during inference of GDR-Net / GDRNPP.}
    \label{fig:runtimeportions}
\end{figure}

GDRNPP \cite{liuShanicelGdrnpp_bop20222024} is a 6D object pose estimation architecture, that constitutes an enhanced version of GDR-Net \cite{wang2021gdr}.
It estimates the pose of an object given an RGB image, by firstly detecting relevant image regions, containing the object region, then predicting relevant features in these image regions using Convolutional Neural Networks (CNN) in the form of a backbone and a subsequent Geo Net, based on which a PnP-Module directly regresses the rotation and translation from the learned features. Subsequently a depth-based pose refinement can be performed as an optional step.\\
We choose to adopt the GDRNPP architecture as a base for the proposed scalable 6D object pose estmation models, due to its high performance in the BOP Challenge \cite{BOPBenchmark6Da} and for its highly adoptable architecture, as demonstrated in a range of diverse extensions \cite{epro, stereo}.
To identify relevant parts of the architecture for inference time optimization, we subdivide the process of GDRNPP into six conceptual stages, i.e., \textit{Data Load}, \textit{Backbone}, \textit{Geo Head}, \textit{Data Process}, \textit{Patch PnP} and \textit{Data Process Afterwards}, which are illustrated with the architecture of GDR-Net (respectively GDRNPP) in Fig. \ref{fig:Architecture of GDRNPP}.
Preliminary experiments show that the major part of inference time are caused by the Data Load, Backbone and Geo Head stages (Fig. \ref{fig:runtimeportions}). Therefore, in the following we propose multiple changes to the architecture in these conceptual stages to reduce inference time, while aiming to preserve accuracy.

\section{Methodology}

In this section, we show potential optimizations to GDRNPP \cite{liuShanicelGdrnpp_bop20222024} especially by choosing suitable backbone architectures (see Sec. \ref{sec:methodology:backbone}) and proposing alterations to the applied Geo head architecture (see Sec. \ref{sec:methodology:geohead}) akin to the proposed optimizations of Pöllabauer et al. \cite{pollabauer2024fast}.
Therefore, we select the five backbones from a pool of 22 candidate backbones which exhibit outstanding performance under a specified GMAC budget (see Sec. \ref{sec:methodology:backbone}).
Furthermore, we propose 11 potential alterations of the Geo Head architecture, from which we chose 4 according to their performance on the LM-O dataset (see Sec. \ref{sec:methodology:geohead}).
The combination of selected backbones and Geo Head architecture results in a pool of 40 candidate architectures.
We propose the AMIS algorithm, that selects a subset of these architectures, which allows efficient inference time and estimate quality trade-off by choosing the appropriate architecture from that pool (see Sec. \ref{sec:methodology:amis}).
In the end, we propose some implementation optimizations regarding the GDRNPP net architecture, that we applied to further increase inference time (see Sec. \ref{sec:methodology:misc}).

\subsection{Backbone}
\label{sec:methodology:backbone}

\begin{figure}[htbp]
    \centering
    \begin{subfigure}[t]{0.44\textwidth}
        \raisebox{-\height}{\includegraphics[width=\textwidth,height=5cm,keepaspectratio]{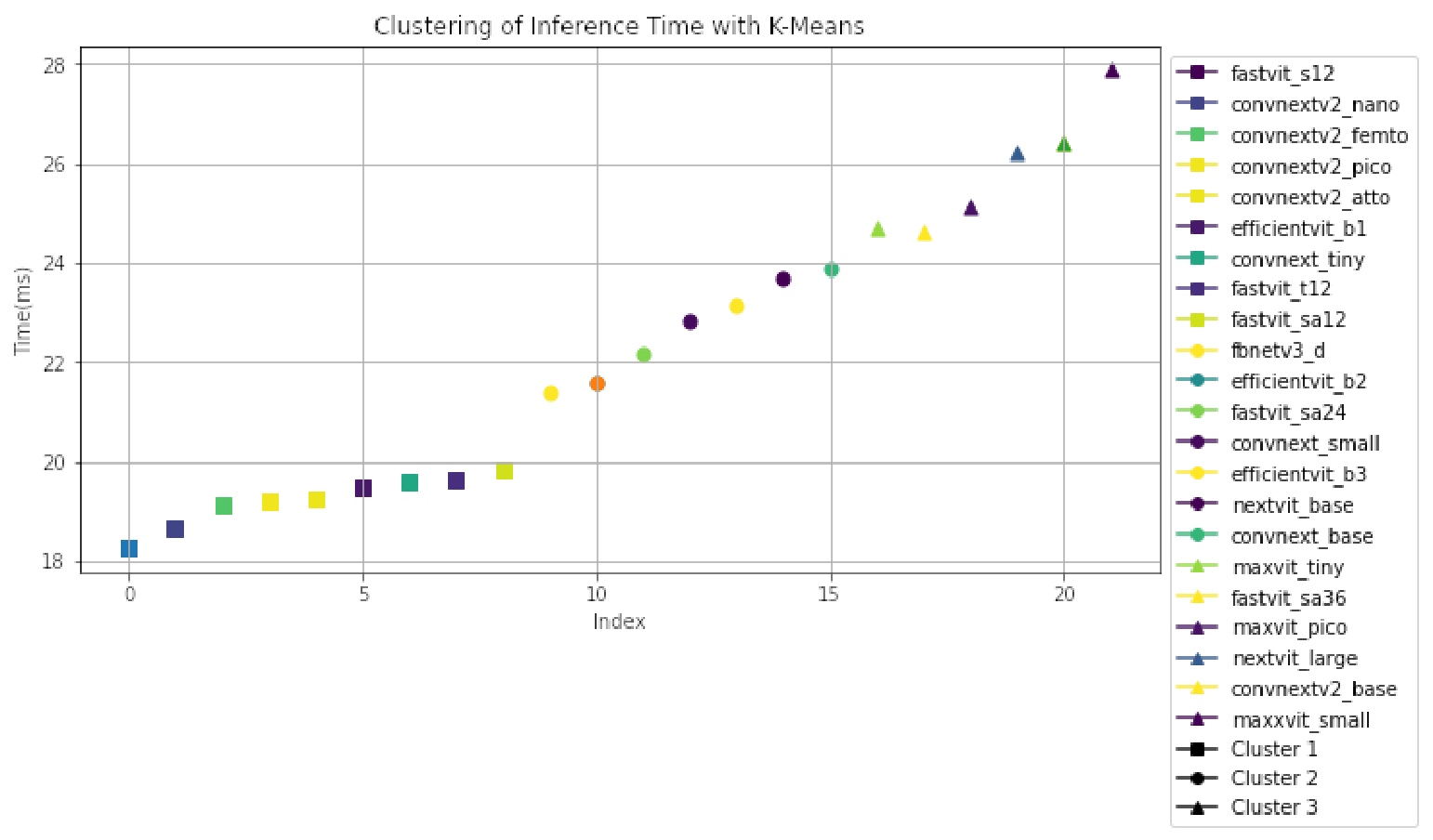}}
        \caption{K-means Clustering of Backbone according to inference time}
        \label{fig:Backbone A}
    \end{subfigure}
    \hfill
    \begin{subfigure}[t]{0.52\textwidth}
        \raisebox{-\height}{\includegraphics[width=\textwidth,height=5cm,keepaspectratio]{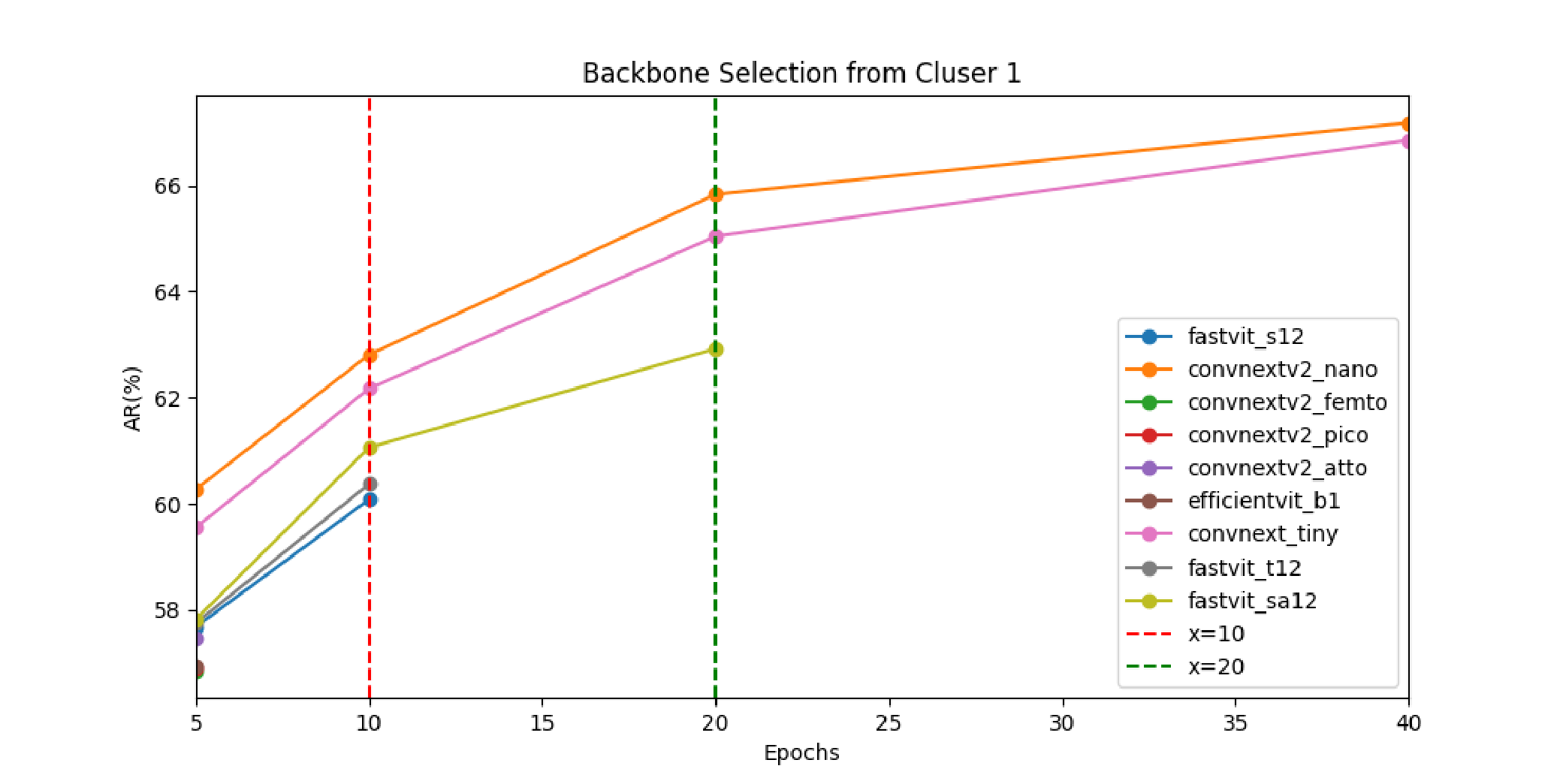}}
        \caption{Successive halving algorithm on Cluster 1}
        \label{fig:Backbone B}
    \end{subfigure}    
    \caption[Backbone Selection]{\raggedright 
    Intermediate results during backbone selection using k-means clustering and the Successive Halving Algorithm with the mean of MSPD, MSSD and VSD as selection criterion.}
    \label{fig:Backbone Selection}
\end{figure}

The impact of choosing different backbones for feature extraction on inference time and accuracy is crucial, prompting experiments with various models available in the Timm \cite{TimmPyTorchImage2024,} package, which offers a broad array of pre-trained CNN and ViT \cite{yuan2021tokens,} models, enabling access to advanced architectures and pre-trained weights which facilitate transfer learning and accelerate model development.\\
To select the most effective models, we set our criteria based on balancing performance speed and accuracy. Faster models with a \textit{Giga Multiply-Accumulate Operations per Second} ratio up to the standard set by GDRNPP were prioritized.
Among models with similar sizes, those offering superior accuracy were chosen. Considering Transformer models' effectiveness in computer vision, variants like Nextvit \cite{li2022next,} and Maxvit \cite{tu2022maxvit,} were selected for their compact sizes suitable for the 256x256 input image size used in our model. This approach ensures a reasonable selection of backbones that optimize both computational efficiency and task performance.\\
We identified 22 candidates for the backbone architecture by analyzing their inference time distribution, which appeared to cluster into three distinct groups, as shown in Fig. \ref{fig:Backbone A} \cite{tu2022maxvit, vasu2023fastvit, woo2023convnext, cai2022efficientvit,}.
Accordingly, we employed the k-means clustering algorithm to categorize these candidates into three groups.
For each group, we utilized the successive halving algorithm to select the best-performing backbone candidate in the group (Fig. \ref{fig:Backbone B}), selecting the five best performing backbones after five training epochs, the best three backbones after ten epochs, until the best performing backbone remained.\\
Notably, we also included Maxvit \cite{tu2022maxvit} for its competitive accuracy and FastVit \cite{vasu2023fastvit} for ts superior speed.
This strategic approach enables a focused evaluation of backbones that potentially optimize both speed and accuracy in our model.

\subsection{Geo Head}
\label{sec:methodology:geohead}
\begin{figure}[htbp]
    \centering
    \begin{subfigure}[b]{0.48\textwidth}
        \centering
        \includegraphics[width=\textwidth,keepaspectratio]{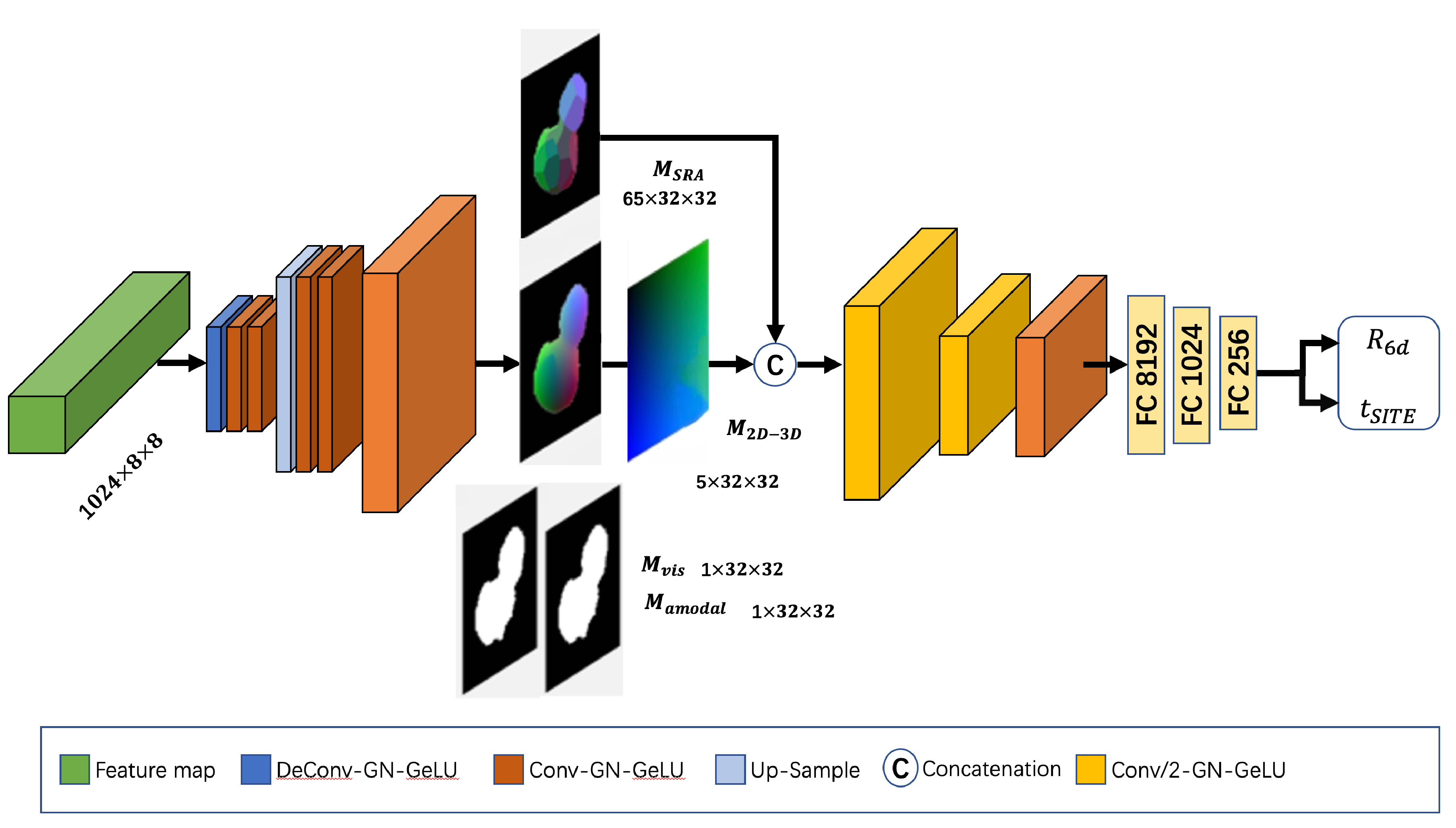}
        \caption{Geo Head variation 1}
        \label{fig:Geo Head variation 1_a}
    \end{subfigure}
    \hfill
    \begin{subfigure}[b]{0.48\textwidth}
        \includegraphics[width=\textwidth,keepaspectratio]{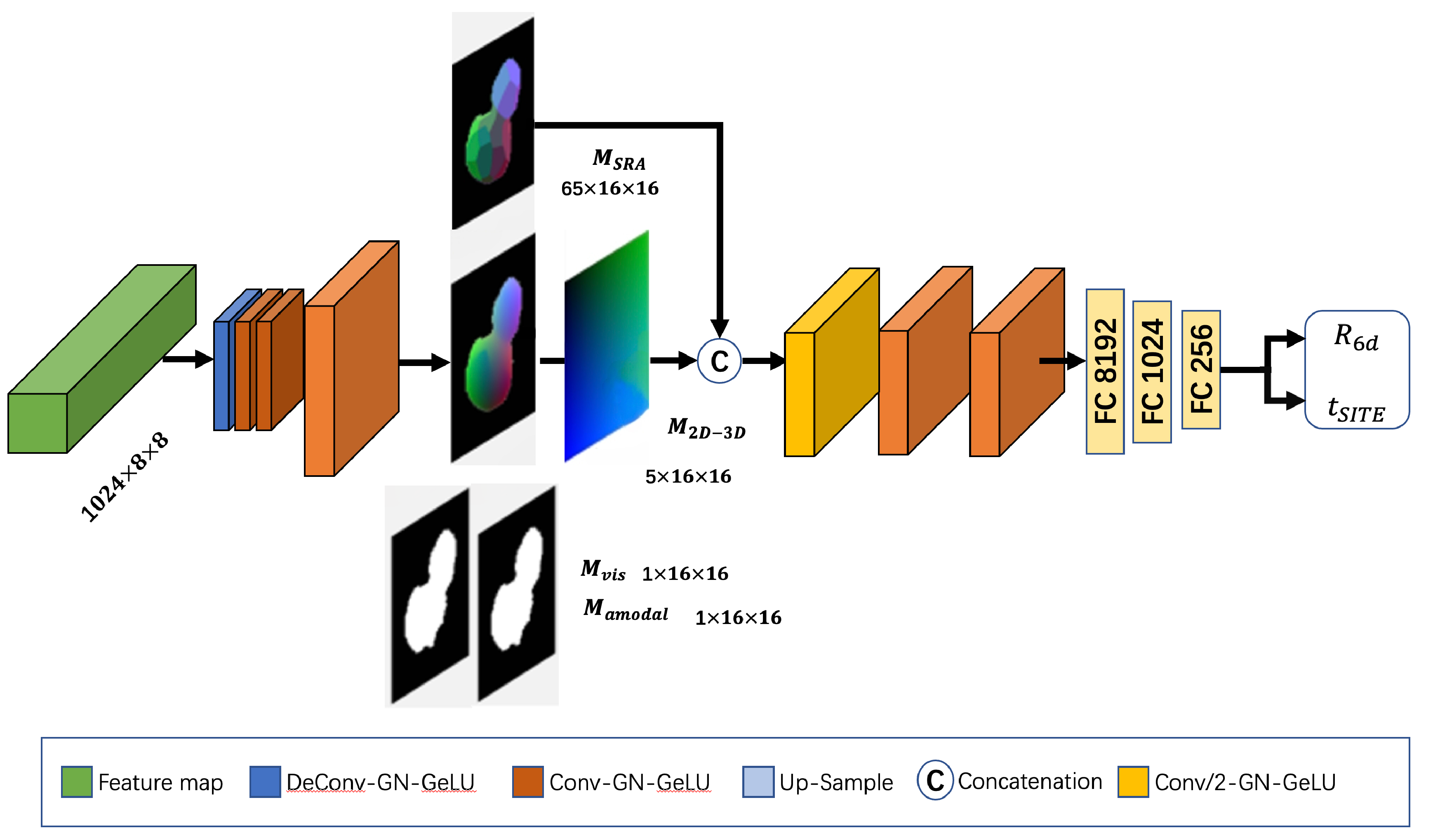}
        \caption{Geo Head variation 2}
        \label{fig:Geo Head variation 2_a}
    \end{subfigure}
    
    \caption[all Geo Head structure]{\raggedright Geo Head Variations GDR-Net architecture with the adaptations Geo Head variation 1 and 2, which reduce the number of up-sample layer of the
     vanilla Geo Head part.}
    \label{fig:all Geo Head structure}
\end{figure}
\begin{figure}[htb]
    \centering
    \begin{subfigure}[b]{0.3\textwidth}
        \includegraphics[width=\textwidth,keepaspectratio]{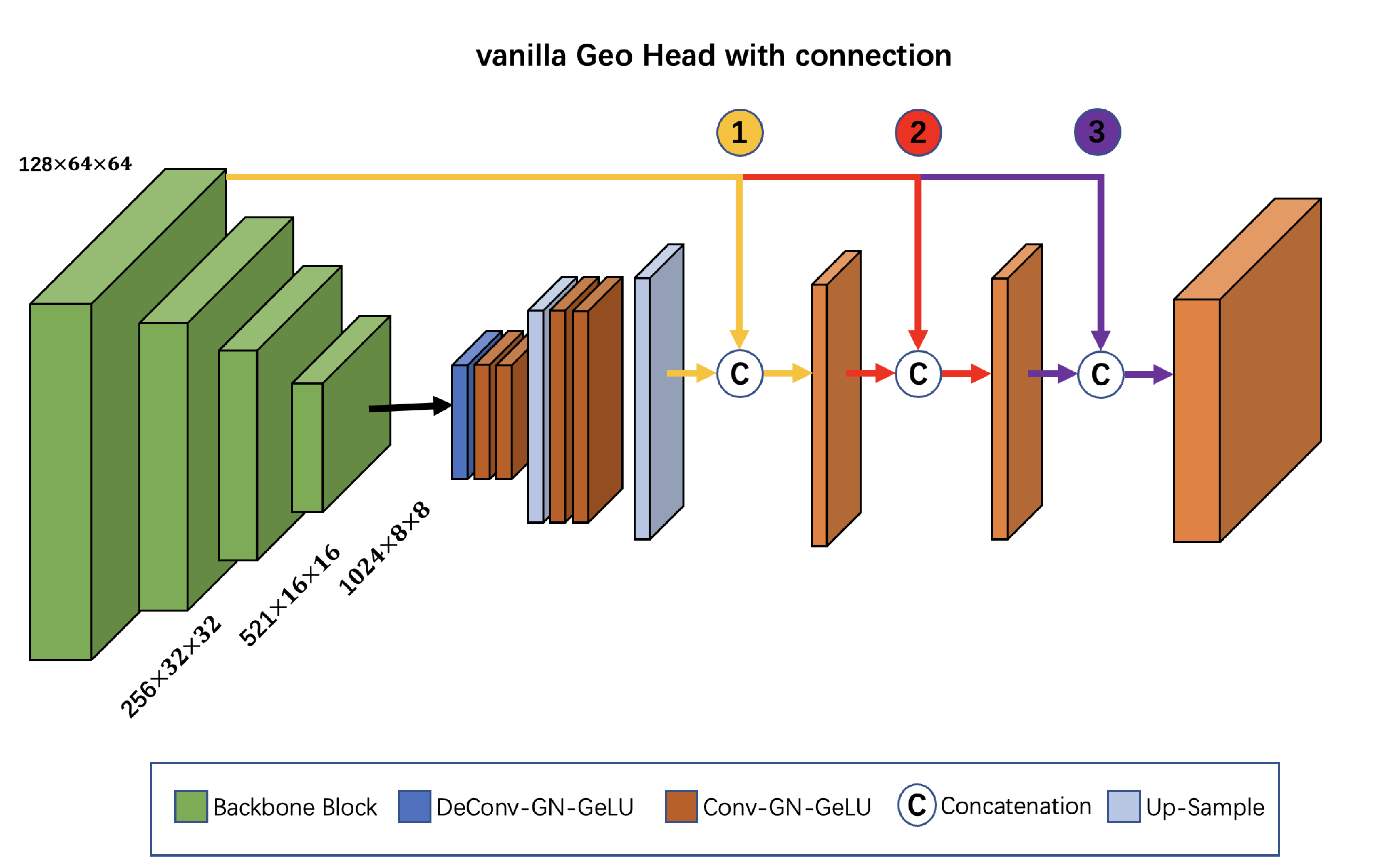}
        \caption{vanilla Geo Head with connection}
        \label{fig:vanilla Geo Head with connection}
    \end{subfigure}
    \begin{subfigure}[b]{0.3\textwidth}
        \includegraphics[width=\textwidth,keepaspectratio]{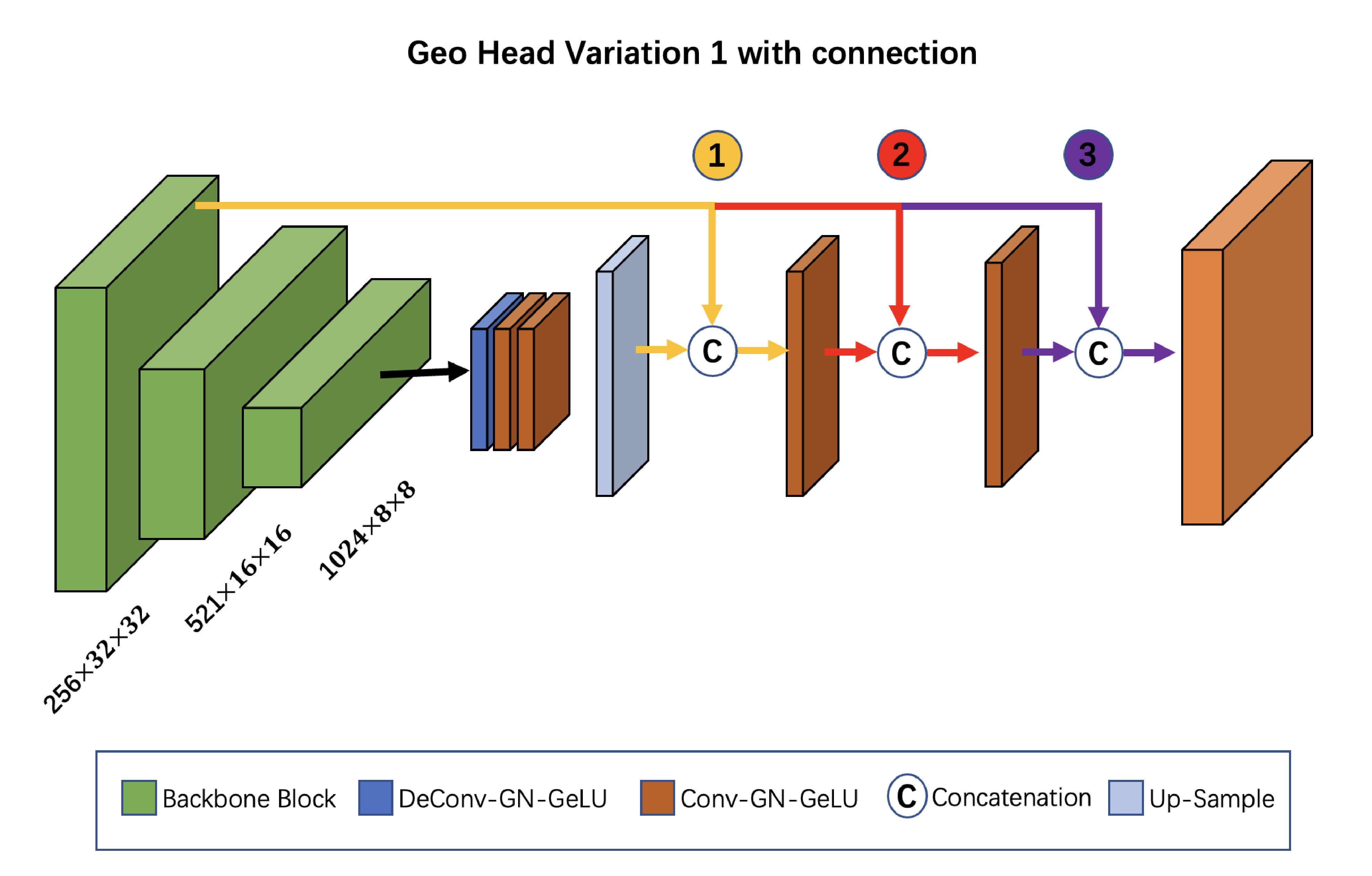}
        \caption{Geo Head Variation 1 with connection}
        \label{fig:Geo Head Variation 1 with connection}
    \end{subfigure}
    \begin{subfigure}[b]{0.3\textwidth}
        \includegraphics[width=\textwidth,keepaspectratio]{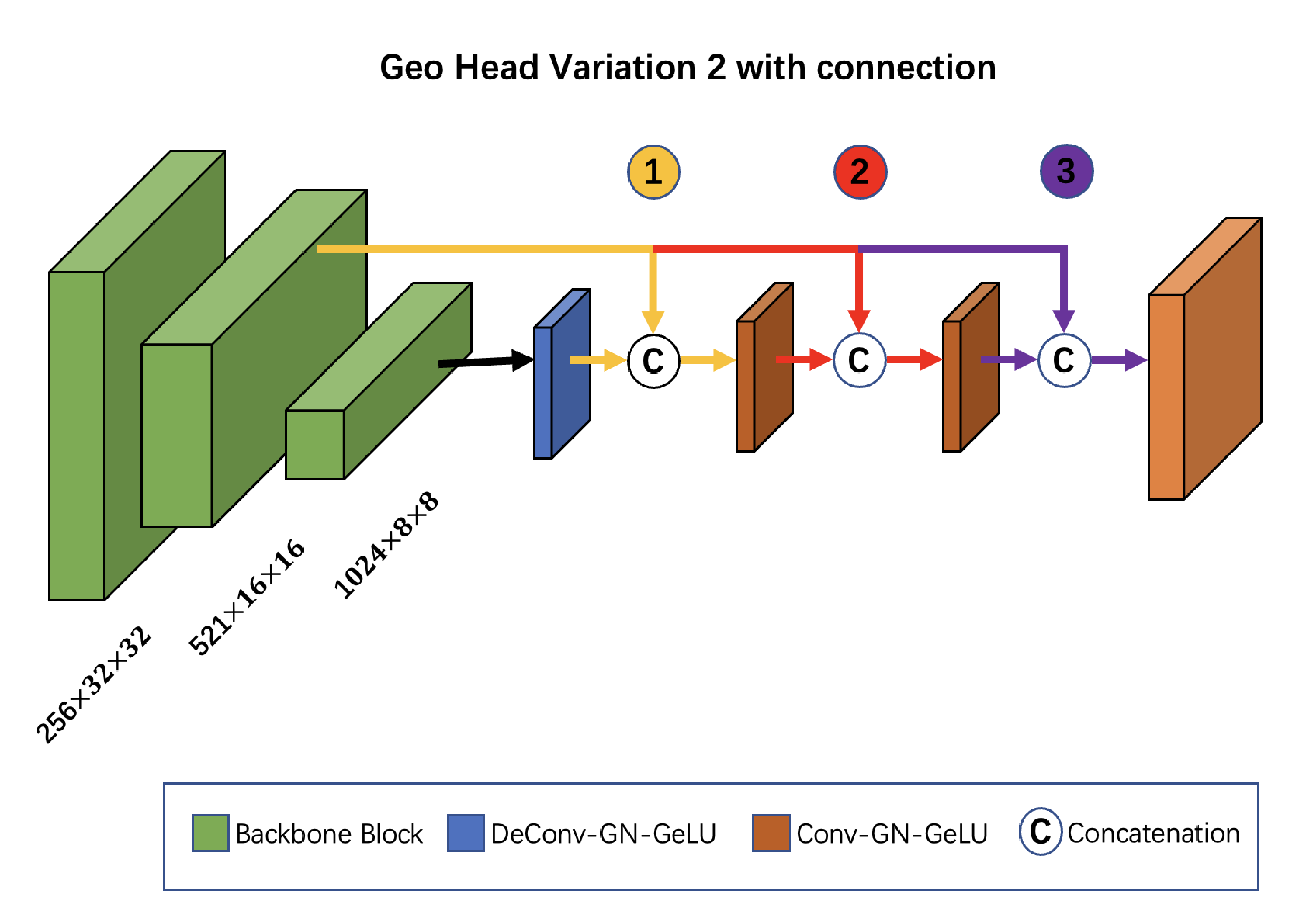}
        \caption{Geo Head Variation 2 with connection}
        \label{fig:Geo Head Variation 2 with connection}
    \end{subfigure}
    \\ 
    \caption[Connection in Geo Head variation 1]{\raggedright Candidates for adding connection in vanilla Geo Head, Geo Head variation 1, Geo Head variation 2. The connection can be added in one of positions labeled \ding{172}, \ding{173} and \ding{174}.}
    \label{fig:Connection in Geo Head variation 1}
\end{figure}

The Geo Head part of GDRNPP is one of the most time consuming parts during inference.
The standard structure of the Geo Head comprises three convolution blocks processing the $1024\times 8 \times 8$ output from the Backbone including up-sampling and convolution layers.
To reduce the inference time requirements, we propose two variants of the vanilla Geo Head architecture: \textit{Geo Head Variation 1}, reducing the convolution blocks to two and modifying the layer setup to handle changes in the feature map size without upsampling, and \textit{Geo Head Variation 2}, which further streamlines this process by eliminating an additional upsampling layer and adjusting the convolution sequence, resulting in smaller feature map sizes and faster data processing.
These modifications are aimed at decreasing the computational load and accelerating inference time by reducing the number of operations the GPU processes and by lowering the input size to the Data Process part, enhancing overall performance.
The proposed architecture variations are illustrated in Fig. \ref{fig:all Geo Head structure}.\\
Inspired by U-Net \cite{ronneberger2015u,}, we implemented skip connections in our neural network architectures to address the gradient vanishing problem and enhance learning capabilities by directly connecting layers across the network. These connections facilitate detail recovery and image segmentation by leveraging rich contextual information during up-sampling. However, introducing skip connections adds complexity to the model, increases computational demands, and may impact the network’s ability to generalize to unseen data. To balance these factors, we simplified the skip connection structure to involve minimal additional computations, ensuring that connected feature maps match in size to avoid unnecessary computations.\\
For each of the three Geo Head Variants (\textit{Vanilla}, \textit{Variation 1} and \textit{Variation 2}), there are three different candidate locations, where skip connections could be added regarding the aforementioned criteria.
These locations are illustrated in Fig. \ref{fig:Connection in Geo Head variation 1}.

\subsection{Adaptive Margin-Dependent Iterative Selection (AMIS) algorithm}
\label{subsec:Scalable 6D Pose Estimation: AMIS}
\label{sec:methodology:amis}
 
\begin{figure}[t]
    \centering
    \begin{subfigure}[b]{0.48\textwidth}                 \includegraphics[width=\textwidth,keepaspectratio]{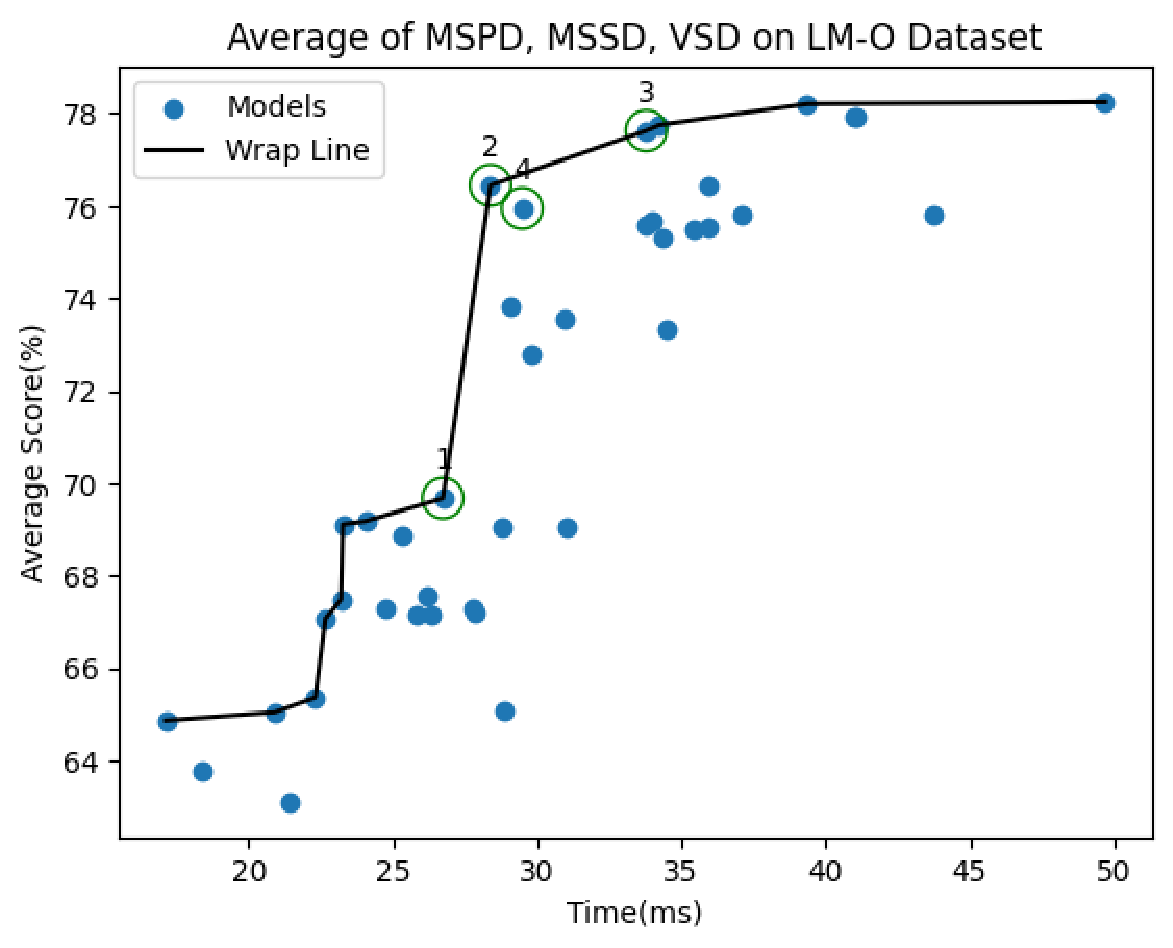}
        \caption{Wrap line showing models with a preferable trade-off between inference time and accuracy.}
        \label{fig:Example for wrap line and sweet point}
    \end{subfigure}
    \hfill
    \begin{subfigure}[b]{0.48\textwidth}
    \includegraphics[width=\textwidth,keepaspectratio]{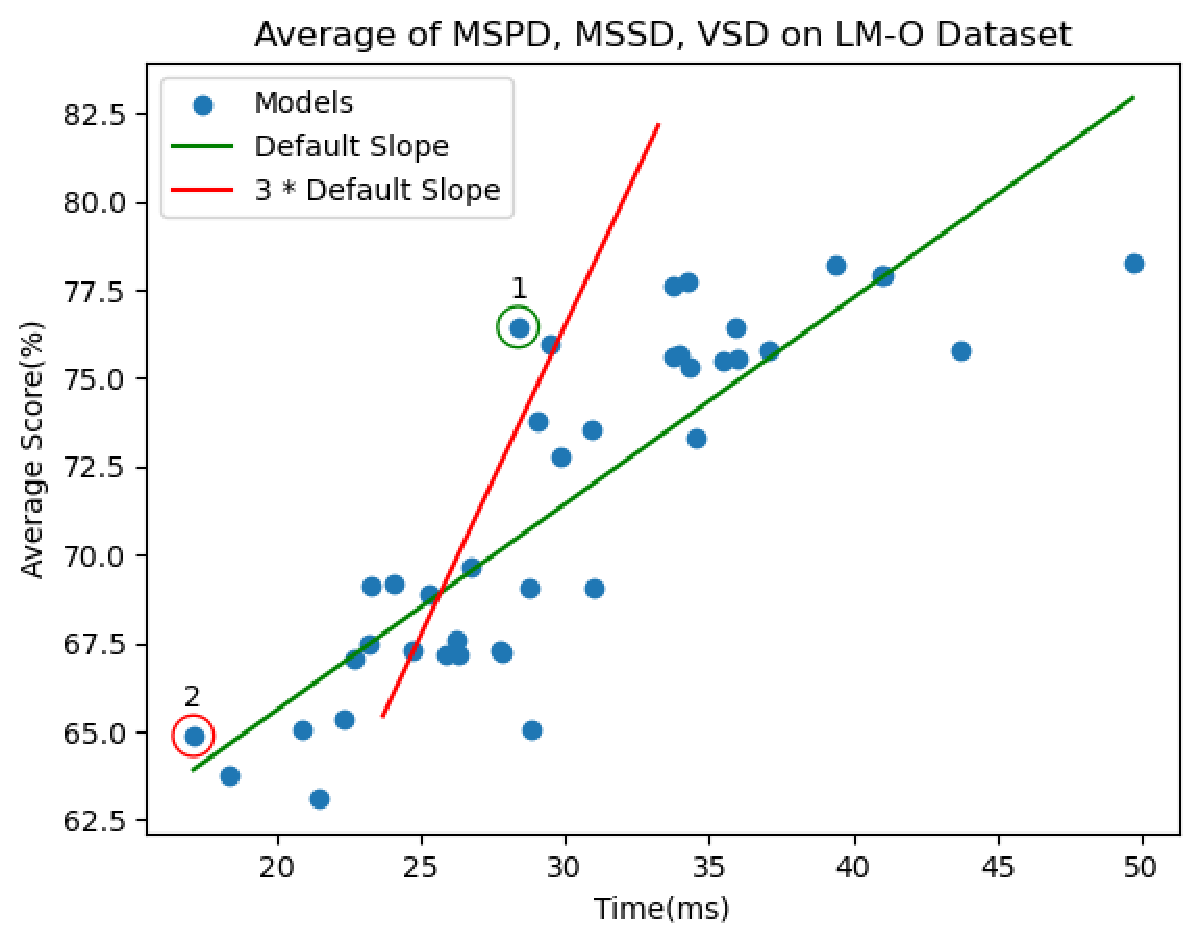}
        \caption[Example for changing slope]{\raggedright \textit{Default slope} and slope with \textit{adjustment factor} in the AMIS algorithm.}
        \label{fig:Example for changing slope}
    \end{subfigure}
    \\
    \caption{Visualization of substages of the AMIS algorithm.}
    \label{fig:Example to explain the theory of AMIS algorithm}
\end{figure}

We propose the Adaptive Margin-Dependent Iterative Selection (AMIS) algorithm to identify a subset of models that excel regarding their estimate quality in comparison to their time budget, i.e., inference time, over arbitrary datasets.
Existing strategies often average results over datasets, which essentially overweights the results of datasets that require high inference time, due to their absolute inference time differences being higher.
In contrast to that, we opt for a strategy that reduces the influence of this effect.
The proposed strategy pinpoints models that are part of an optimal wrap-line (Fig. \ref{fig:Example for wrap line and sweet point}) in the space of inference time and estimation quality, which show substantial accuracy improvement over its quicker counterparts with only minimal gains compared to slower successors.
Recognizing these models aids in selecting models that effectively balance speed and accuracy, enhancing model selection strategies.\\
In an initial phase, we measure inference time and accuracy for each candidate model and each dataset.
For each dataset we fit a straight line in the 2D space spanned by inference time and accuracy metrics using linear regression, which we call the \textit{default slope}.
For each model and dataset, we calculate the distance of its result from this default slope, normalizing them for each dataset on a scale of $0$ to $100$.
These scores are weighted depending on the desired dataset weight, resulting in a final score for each model.
We rank these scores assigning every model a fixed amount of scoring points depending on their rank.
We then iteratively repeat that ranking process for 100 \textit{adjustment factors}, between 0.001 and 3, which are multiplied with the \textit{default slope} (Fig. \ref{fig:Example for changing slope}).
We accumulate the ranking points per model for different adjustment factors except from the case, where the 10 best performing models do not change in comparison to the previous adjustment factor.
This procedure ensures, that the resulting model selection is robust against diverse trade-off preferences. After completing this step, we select the best ranking model repeating the process with the repeating models, until the number of selected models meets the task-specific requirements.

\subsection{Other optimization}
\label{sec:methodology:misc}
To reduce extended inference times linked with dynamic tensor creation, which is resource-intensive due to repeated memory allocation and initialization, we have adopted alternative measures.
Our strategy minimizes new tensor creation by pre-allocating tensor memory before inference, using a reusable tensor pool throughout the inference cycle to avoid frequent new allocations.
This approach enhances efficiency, which is particularly critical in real-time applications.
\section{Results}

In this section, we provide a quantitative analysis of the proposed Geo Head architecture improvements (see Sec. \ref{sec:results:preliminary}), proposing 40 candidate architectures based on the best performing ones.
Furthermore, we show the results of the models selected by the proposed AMIS for multiple datasets (see Sec. \ref{sec:results:amis}), showing that the selected models outperform GDRNPP and provide a reasonable model selection regarding a desired trade-off between inference time and accuracy on multiple datasets.

\subsection{Preliminary Experiments}
\label{sec:results:preliminary}

In sections \ref{sec:methodology:geohead} and \ref{sec:methodology:misc} we proposed alterations to the Geo Head architecture of GDRNPP with the goal of reducing inference time while maintaining a high level of accuracy.
We evaluate the performance of GDRNPP using these Geo Head architectures on the LM-O dataset.
The results of this evaluation are illustrated in Tab. \ref{tab:Comparison of Different Modifications on LM-O}.
These results indicate, that the proposed general optimizations \textit{B0} and \textit{C0} positively influence the inference time, while maintaining the desired level of precision, measured by MSPD, MSSD, VSD and AR, which give insight into a wide variety of aspects of 6D object pose estimation provided by the BOP Toolkit\cite{hodan2024bop}.
Furthermore, the results indicate, that the introduced variations of the Geo Head architecture reduce the inference time even further.
We ablate the influence of architectural changes in more detail in section \ref{sec:results:geohead}.\\
Based on the results, we chose the Geo Head architectures \textit{C0}, \textit{E0}, \textit{F0}, and \textit{F2} due to their high performance in comparison to the required inference time.
We combine those architectures with the previously identified candidate backbones to obtain 20 combinations.
Those candidates are assigned with a number for their identification according to Tab. \ref{tab:Encode Experiment Candidates}.
We evaluate each of these architecture combinations with and without the optional refinement step of GDR-NET in the subsequent experiments, essentially resulting in 40 candidate architectures.

\begin{table}[H]
\centering
\caption[Comparison of Different Modifications on LM-O]{\raggedright
Quantitative results of the proposed Geo Head candidates ond the LM-O dataset regarding MSPD, MSSD, VSD, AR and Inference time. The Geo Head architectures that we chose to construct candidate architecture are indicated in \textbf{bold}.}
\label{tab:Comparison of Different Modifications on LM-O}
\begin{tabular}{@{}l|l|c|c|c|c|c@{}}
\toprule
\multirow{2}{*}{\textbf{Row}} & \multirow{2}{*}{\textbf{Method}} & \textbf{MSPD} & \textbf{MSSD} & \textbf{VSD} & \textbf{AR} & \textbf{Time} \\
 &  &  \textbf{\%} & \textbf{\%} & \textbf{\%} & \textbf{\%} & \textbf{ms}\\
\midrule
A0 & GDRNPP & 87.14 & 67.03 & 52.38 & 68.85 & 28.8 \\
\midrule
B0 & A0: Optimizations & 87.14 & 67.07 & 52.38 & 68.86 & 26.72 \\
\midrule
\textbf{C0} & \textbf{B0: Data Process Optimization} & 87.14 & 67.03 & 52.38 & 68.86 & \textbf{25.29} \\
\midrule
D0 & C0: Add connection in location 1 & \textbf{87.66} & \textbf{67.29} & \textbf{52.53} & \textbf{69.16} & 28.79 \\
D1 & C0: Add connection in location 2 & 86.89 & 66.75 & 51.91 & 68.82 & 29.17 \\
D2 & C0: Add connection in location 3 & 87.39 & 66.64 & 52.01 & 68.68 & 25.68 \\
\midrule
\textbf{E0} & \textbf{C0: Vanilla Geo Head$\rightarrow$variation 1} & \textbf{87.11} & \textbf{67.78} & \textbf{52.58} & \textbf{69.15} & \textbf{24.05} \\
E1 & E0: Add connection in location 1 & 86.71 & 67.2 & 52.11 & 68.7 & 24.47 \\
E2 & E0: Add connection in location 2 & 87.16& 67.13& 52.42& 68.9 & 24.32 \\
E3 & E0: Add connection in location 3 & 87.30& 66.83& 51.90 & 68.68 & 24.13 \\
\midrule
\textbf{F0} & \textbf{C0: Vanilla Geo Head$\rightarrow$variation 2} & 85.23 & 65.40 & 50.61 & 67.08 & 22.66 \\
F1 & F0: Add connection in location 1 & 85.29& 65.72& 50.88 & 67.3 & 22.22 \\
\textbf{F2} & \textbf{F0: Add connection in location 2} & \textbf{84.96}& \textbf{66.26}& \textbf{51.27}& \textbf{67.49} & 23.20 \\
F3 & F0: Add connection in location 3 & 81.46& 58.30& 44.82 & 61.53 & 23.11 \\
\midrule
G0 & F2: convnext\_base$\rightarrow$convnextv2\_nano & 83.47 & 63.00 & 48.15 & 64.87 & \textbf{17.12} \\
\midrule
H0 & E0: convnext\_base$\rightarrow$convnextv2\_base & 87.31 & 67.62 & 52.63 & \textbf{69.18} & 24.05 \\
\bottomrule
\end{tabular}
\end{table}

\begin{table}[H]
\centering
\caption[Coding Experiment Candidates]{\raggedright Candidate Models consisting of one of the previously identified backbones and adapted Geo Head candidates (F0, F1, E0, E1).} 
\label{tab:Encode Experiment Candidates}
\begin{tabular}{@{}l|cccc@{}}
\toprule
\multirow{2}{*}{\textbf{Backbone}} & \multicolumn{4}{c}{\textbf{configuration}} \\
 & F0 & F2 & E0 & C0 \\
\midrule
fastvit\_s12 & 1 & 2 & 3 & 4 \\
convnextv2\_nano & 5 & 6 & 7 & 8 \\
convnext\_base & 9 & 10 & 11 & 12 \\
convnextv2\_base & 13 & 14 & 15 & 16 \\
maxxvit\_small & 17 & 18 & 19 & 20 \\
\bottomrule
\end{tabular}
\end{table}

\subsection{Geo Head}
\label{sec:results:geohead}
In this section, we summarize the findings from the proposed alterations to the Geo Head part of GDR-Net / GDRNPP as evaluated on the LM-O dataset (see Tab. \ref{tab:Comparison of Different Modifications on LM-O}).\\
Our results indicate that the proposed optimizations (see Sec. \ref{sec:methodology:misc}) improve the inference time without a relevant impact on the measured accuracy.\\
The proposed Geo Head variation 1 leads to a speed-up in inference time and an improvement in accuracy, constituting it a successful optimization to the Geo Head architecture.
In contrast to that, the proposed Geo Head variation 2, further increased inference time, while reducing accuracy significantly. The availability of such an adaption, however, is highly beneficial when looking for highly accurate 6D pose estimation models under varying inference time budgets.\\
Adding connections within the Geo Head was generally found to negatively impact inference speed, particularly in the vanilla Geo Head setup, where it was deemed not worthwhile due to significant slowdowns without notable accuracy benefits. However, in the second variation of Geo Head, adding a connection at the first location improved both speed and accuracy, making it a promising adjustment for balancing performance metrics.

\begin{figure}[b]   
    \includegraphics[width=\textwidth,keepaspectratio]{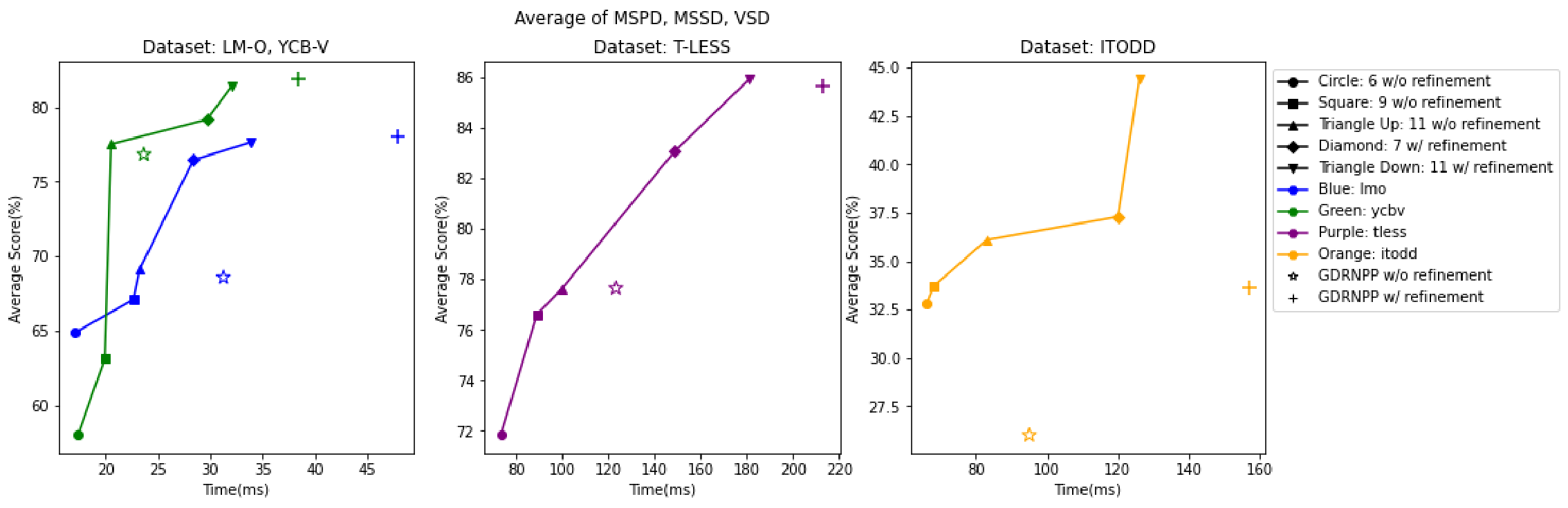}
    \caption[Average of MSPD, MSSD, VSD]{Average over MSPD, MSSD, VSD of the 5 candidate architectures with desireable trade-off between inference time and accuracy using the AMIS algorithm on LM-O (left), YCB-V(left), T-LESS (center) and ITODD (right) dataset.}
    \label{fig:Average of MSPD, MSSD, VSD_reference}
\end{figure}

\subsection{AMIS}
\label{sec:results:amis}

We employ the aforementioned AMIS algorithm (see Sec. \ref{sec:methodology:amis}) to identify a suitable subset of models within the previously identified candidate architectures on the IMO, LM-O, YCB-V, T-LESS, and ITODD datasets \cite{xiang2017posecnn, drost2017introducing, hodan2017t,}.
The results of the five identified candidates are illustrated in Fig. \ref{fig:Average of MSPD, MSSD, VSD_reference} on scatterplot showing inference time and 6D object pose estimation accuracy.
Furthermore, the average results of the models are illustrated in Tab. \ref{tab:Result of selected Candidates vs GDRNPP}.\\
The experimental results show, that even with minimal time budget, i.e., when we expect our model to perform the fastest inference, the required inference time is reduced by 35\% in comparison to GDRNPP, while the achieved performance only drops by 3\% measured by the average of MSPD, MSSD, and VSD.
As the time budget increases, the performance of our candidates also gradually improves. This variation can adapt to the complex scenarios of different time and accuracy requirements in industrial environments. It is noteworthy that compared to GDRNPP, using about 31\% additional time can lead to approximately a 25\% improvement in performance.
Fig. \ref{fig:Average of MSPD, MSSD, VSD_reference} furthermore shows that the selected models show an increase in accuracy with increased inference time for all 4 datasets.

\begin{table}[ht]
\centering
\caption{Quantitative results of the candidate architectures identified by the AMIS algorithm, showing that they constitute a desirable trade-off between inference time and accuracy.}
\label{tab:Result of selected Candidates vs GDRNPP}
\begin{tabular}{@{}l|cccc@{}}
\toprule
\multicolumn{5}{c}{\textbf{5 selected Candidates vs GDRNPP}}\\
\midrule
\multirow{2}{*}{\textbf{Candidates}}& \textbf{Average of} & \multirow{3}{*}{\textbf{ADD}}& \textbf{Average of }&  \multirow{3}{*}{\textbf{Time}}\\
& \textbf{MSPD, MSSD} &  & \textbf{MSPD, MSSD,} &   \\
\multirow{2}{*}{\textbf{Number}}& \textbf{VSD} &  & \textbf{VSD, ADD} &   \\
& \% & \% & \% & \%  \\
\midrule
6 without ref. & -2.82 & -16.62 & -3.42 & -35.71\\
\midrule
9 without ref. & +2.06 & -9.82 & +1.66 & -24.86\\
\midrule
11 without ref.{ } & +10.12 & +3.0 & +10.44 & -17.61\\
\midrule
7 with ref. & +16.24 & +41.28 & +20.56 & +15.81\\
\midrule
11 with ref. & +25.14 & +50.83 & +30.35 & +30.74\\
\bottomrule
\end{tabular}
\end{table}
\section{Conclusion}
We presented a fast and scalable pose estimator, dynamically adjusting to custom needs of estimation quality versus inference times trade-offs.
To that end, we proposed 40 candidate architectures which aim to otimize the trade-off between inference time and 6D object pose estimation accuracy based on GDRNPP, which were curated by choosing promising backbones and identifying beneficial architectural changes to the Geo Head part of the GDRNPP architecture.
Additionally, we proposed the AMIS algorithm, a tool designed to quantitatively identify model architectures that represent sweet spots which effectively address the challenges posed by the diverse scales of precision and time across various datasets. 
In our comparison with GDRNPP, we demonstrate the candidates selected by the AMIS algorithm exhibit an outstanding inference time and progressively enhancing model accuracy for a wide variety of inference time budgets.\\
Future directions might include temporal trajectory composition of different frequencies for long-horizon tasks, policy composition methods on large-scale datasets, and policy distillation from composed policies.
For further research, there are some potential areas of improvement: adopting end-to-end methods that integrate detection or segmentation and exploring other lightweight methods like teacher-student models.

\clearpage
\printbibliography
\end{document}